\documentclass{article}

\usepackage[preprint]{corl_2026} % Uncomment for pre-prints (e.g., arxiv); This is like ``final'', but will remove the CORL footnote.

\usepackage[utf8]{inputenc} % allow utf-8 input
\usepackage[T1]{fontenc}    % use 8-bit T1 fonts
\usepackage{hyperref}       % hyperlinks
\usepackage{url}            % simple URL typesetting
\usepackage{booktabs}       % professional-quality tables
\usepackage{graphicx}        % graphics
\usepackage{adjustbox}       % max width table scaling
\usepackage{amsfonts}       % blackboard math symbols
\usepackage{amsmath,amssymb} % math equations and symbols
\usepackage{nicefrac}       % compact symbols for 1/2, etc.
\usepackage{microtype}      % microtypography
\usepackage{xcolor}         % colors
\usepackage{tikz}
\usepackage{pgfplots}
\pgfplotsset{compat=1.18}
\usetikzlibrary{shapes,arrows,positioning,calc}
\usepackage[capitalize,noabbrev]{cleveref}

\newcommand{\modelname}{\textsc{Parvulus}}
\newcommand{\modelnameshort}{\textsc{Parvulus}}

\title{\modelname{}: A 10M-Scale World Action Model for Robot Manipulation}

\author{
  Abdullah Ali Sivas\thanks{Corresponding Author, ORCID: \href{https://orcid.org/0000-0002-5263-1889}{0000-0002-5263-1889}, Project Page: \href{https://github.com/aasivas/Parvulus}{https://github.com/aasivas/Parvulus} and \href{https://aasivas.github.io/Parvulus}{https://aasivas.github.io/Parvulus}, Model Page: \href{https://huggingface.co/aasivas/Parvulus}{https://huggingface.co/aasivas/Parvulus}} \\
  Huawei Technologies Canada \\
  \texttt{abdullah.ali.sivas@huawei.com} \\
  \And
  Peter Cho-Ho Lam \\
  Huawei Technologies Canada \\
  \texttt{cho.ho.lam@huawei.com} \\
  \And
  Zirui Zhou \\
  Huawei Technologies Canada \\
  \texttt{zirui.zhou@huawei.com}
  \And
  Yong Zhang \\
  Huawei Technologies Canada \\
  \texttt{yong.zhang3@huawei.com}
}

\begin{document}
\maketitle

%===============================================================================

\begin{abstract}
Vision-based robot manipulation increasingly relies on multi-billion parameter models. However, standard imitation learning uses capacity inefficiently. In this work, we show that latent physical grounding improves policy representations. Auxiliary generative decoders ground representations in spatial geometry and dynamics during training. These decoders detach entirely at inference. Trained from scratch on the LIBERO corpus, our 9.9M parameter policy, \modelname{}, achieves a 90-92\% success rate. It operates at $>100\,\text{Hz}$ with $<10\,\text{ms}$ latency. By matching models over $100\times$ larger, \modelname{} advances the Pareto frontier of parameter efficiency among language-conditioned policies.
\end{abstract}


%===============================================================================

\section{Introduction}
\label{sec:introduction}

Recent robot manipulation research focuses heavily on multi-billion parameter models. Modern Vision-Language-Action (VLA) architectures and World Action Models (WAMs) typically scale from 3B to 7B parameters~\citep{kim2024openvla,black2024pi0,cogact2024,yuan2026fastwam,dreamwam2026,yan2026flexpi}. However, scaling model size without physical inductive biases yields diminishing returns. These architectures require substantial compute across training and deployment. Inference latencies typically range from 190 to 800\,ms. While large backbones capture broad semantics, fine manipulation depends on spatial precision and transition dynamics. Consequently, closed-loop control benefits from high decision frequencies ($>100\,\text{Hz}$) and compact footprints.

Standard imitation learning utilizes model capacity inefficiently, forcing scaled architectures to rely on web pretraining to absorb physical dynamics incidentally. In this work, we argue that physical inductive biases must be instilled explicitly into the latent representation. By grounding representations in predictive dynamics during training, we maximize capacity utilization (ensuring each weight internalizes actionable physical knowledge) while preserving the high decision frequencies demanded by closed-loop control. On the 40-task LIBERO benchmark~\citep{liu2023libero}, our 9.9M parameter policy matches models over $100\times$ larger, proving that structured latent grounding, rather than raw scale, drives manipulation competence.

Our framework unites two philosophical stances with two design constraints:

\textbf{Philosophical Stance 1: Latent Physical Grounding.} Robot learning benefits from internalizing dynamics in latent space. We train our backbone under auxiliary predictive objectives. Dynamic frame differencing isolates coordinated motion between the arm and objects. Forward video flow models transition causality. Self-reconstruction preserves spatial scene geometry.

\textbf{Philosophical Stance 2: Data Engineering over Loss Engineering.} Rather than engineering complex loss penalties~\citep{physisforcing2026}, we rely on data engineering. Multi-task generative decoders and noise corruptions implicitly regularize the shared representation against visual shifts.

\textbf{Design Constraint 1: Specialist Mastery from Modest Data.} We train strictly on the modest LIBERO corpus~\citep{liu2023libero,pi_libero_dataset} (1,693 demonstrations, 333k transitions), extracting auxiliary targets directly from demonstrations. We avoid internet-scale pretraining to test whether structured representation learning is sufficient for reliable robot control (defined loosely as $>90\%$ success; while proper reliability demands $>99.9\%$, no current policy attains this).

\textbf{Design Constraint 2: High-Frequency Asymmetric Deployment.} Auxiliary generative decoders must be removable by design. We detach all auxiliary decoders during inference. This maximizes control frequency ($>100\,\text{Hz}$) and keeps memory under 20\,MB. Cross-attention KV-caching decouples perception passes from ODE integration steps.

\begin{figure*}[t]
\centering
\includegraphics[width=\textwidth]{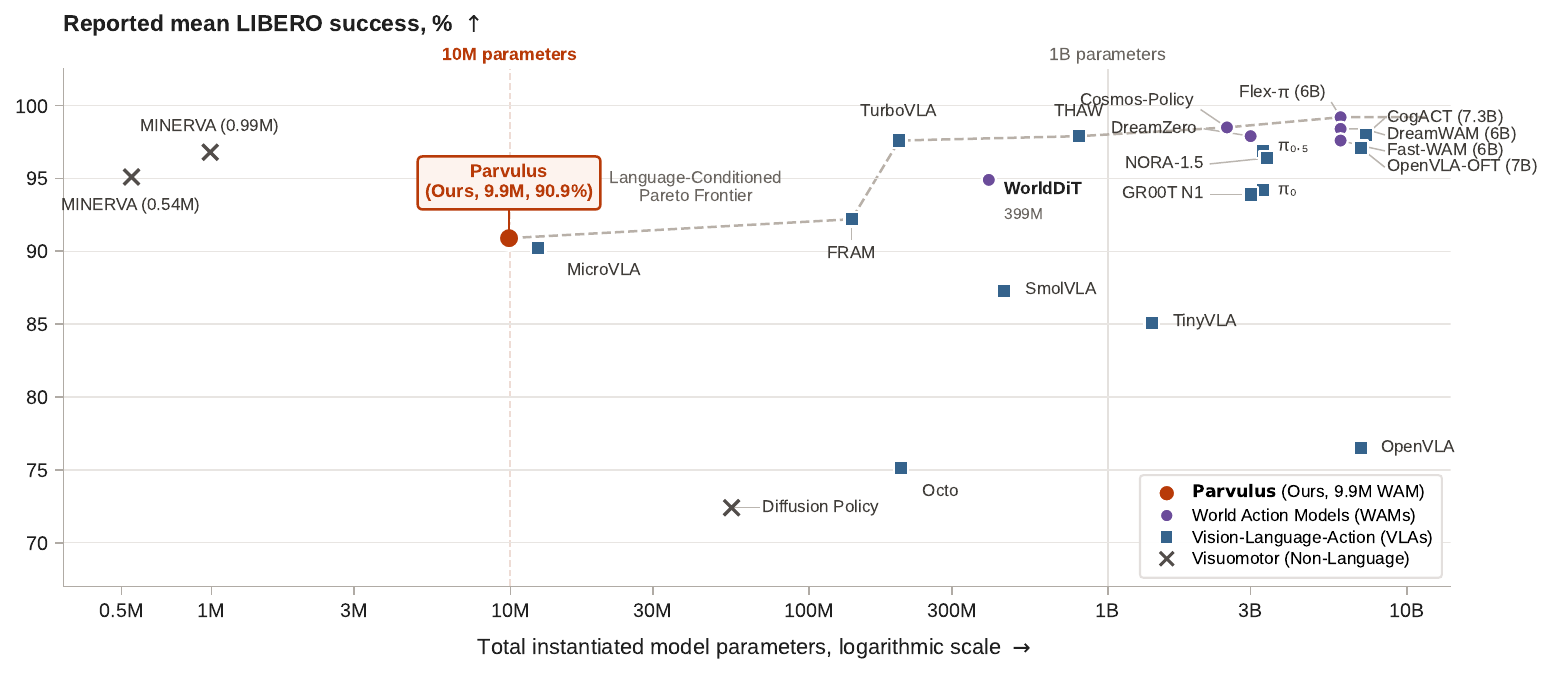}
\vskip -0.05in
\caption{Reported LIBERO manipulation success against total model parameters ($\log_{10}$ scale) across 23 representative architectures. The dashed line traces the language-conditioned Pareto efficiency frontier (connecting \modelnameshort{}, FRAM, TurboVLA, THAW, Cosmos-Policy, and Flex-$\pi$). \modelnameshort{} establishes a new efficiency frontier at 9.9M scale, matching models over $100\times$ larger.}
\label{fig:libero_landscape}
\vskip -0.1in
\end{figure*}

\Cref{fig:libero_landscape} summarizes the LIBERO efficiency landscape across model scales. \modelname{} establishes a new Pareto efficiency frontier among language-conditioned architectures. It achieves 90-92\% success across all four suites. It matches models over 100$\times$ larger while requiring only 9.9M parameters.

\textbf{Contributions.}
Our primary contribution is a training regimen for latent physical grounding. We combine multi-view mixing, multi-scale pooling, and data engineering. Auxiliary decoders regularize latent features during training and detach at test time. Cross-attention KV-caching decouples perception FLOPs from integration steps. Consequently, our standalone 9.9M parameter policy achieves 90-92\% success across 40 LIBERO tasks. It runs at $<10\,\text{ms}$ latency ($>100\,\text{Hz}$) within 20\,MB of memory.

\subsection{Related Work}
\label{sec:related_work}

\textbf{Vision-Language-Action and World Action Models.}
Canonical VLA models adapt large vision-language backbones (3B to 7B parameters) for robot control~\citep{kim2024openvla,black2024pi0,cogact2024,pi05_2025}. However, autoregressive decoding and continuous diffusion heads incur 200 to 800\,ms latency. Recently, World Action Models (WAMs) show that predictive video objectives enrich latent representations during training~\citep{worlddit2026,yuan2026fastwam,dreamwam2026,dreamzero2026,cosmos_policy2026,yan2026flexpi}. Fast-WAM~\citep{yuan2026fastwam} and Flex-$\pi$~\citep{yan2026flexpi} demonstrate the benefit of training-time visual grounding. However, they rely on multi-billion parameter backbones and robot pretraining. WorldDiT~\citep{worlddit2026} also requires pretraining and separate per-suite fine-tuning. In contrast, our approach realizes latent physical grounding in a unified 9.9M policy trained from scratch.

\textbf{Compact Policies and Data Engineering.}
Recent studies investigate compact manipulation architectures~\citep{sendai2026minerva,kim2026microvla,turbovla2026,ito2026fram,dao2026thaw}. MINERVA~\citep{sendai2026minerva} demonstrates that sub-1M parameter CNNs can memorize demonstration paths. However, MINERVA replaces language with a static 40-way task-ID lookup table and degrades under visual shifts. TurboVLA~\citep{turbovla2026} operates at 200M scale, FRAM~\citep{ito2026fram} prunes trajectory-guided visual tokens at 138.7M, and THAW~\citep{dao2026thaw} distills world-model features into a 0.8B student VLA, while MicroVLA~\citep{kim2026microvla} combines existing models and trained action head into 12.4M parameters model with closed weights. Instead of engineering complex physical loss penalties~\citep{physisforcing2026}, we prioritize data engineering. Noise corruptions, cross-view mixing, and detachable decoders implicitly structure the latent space without external perception teachers.

\section{Methodology}
\label{sec:methodology}

\begin{figure*}[t]
\centering
\includegraphics[width=\textwidth]{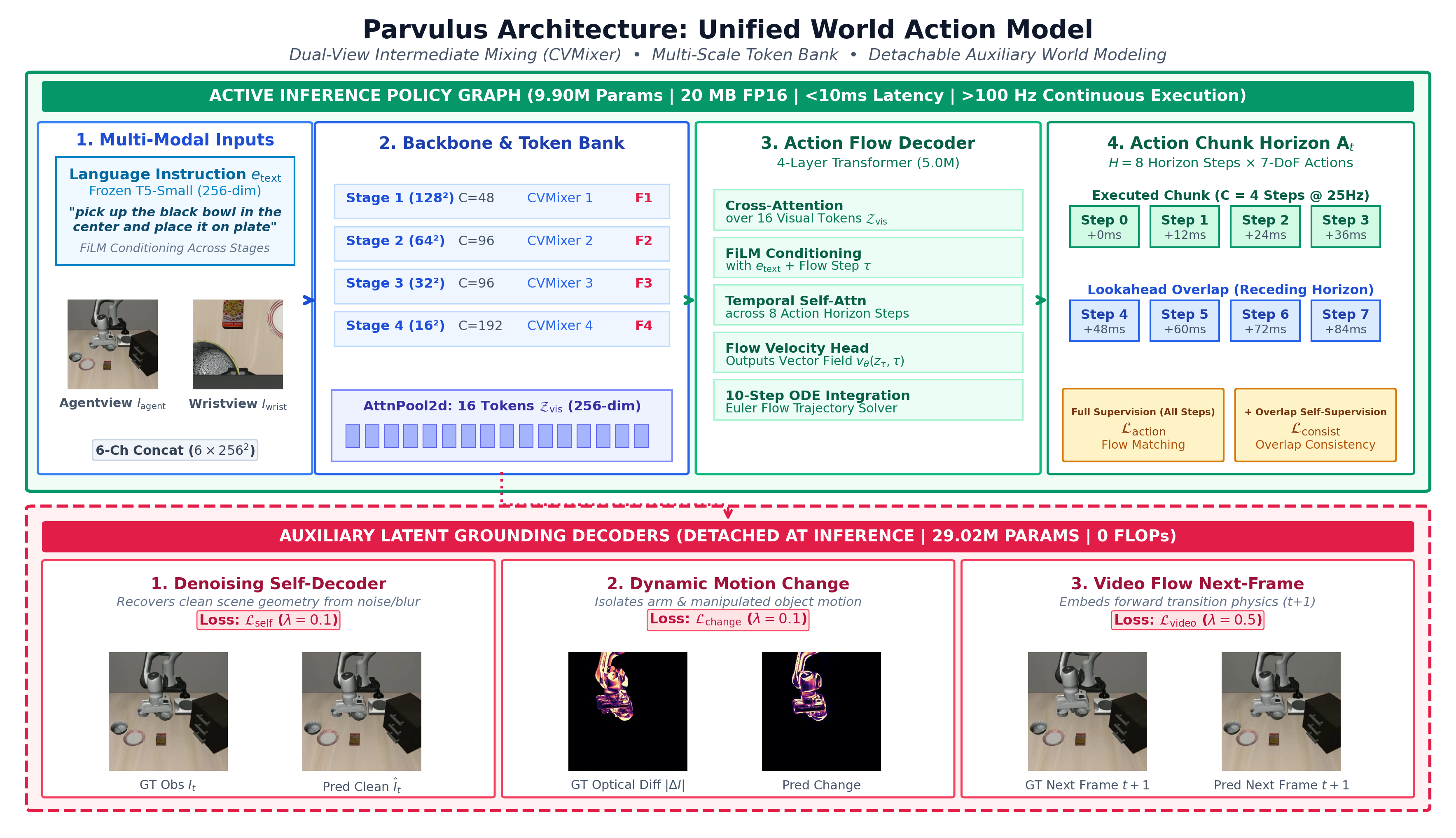}
\caption{\textbf{Overview of the \modelname{} Architecture and Multi-Decoder Auxiliary Grounding.} 
\textit{Top (Active Inference Policy Graph):} Left-to-right inference pipeline ($>100\,\text{Hz}$, $<10\,\text{ms}$, 9.90M parameters). Dual-perspective observations ($I_{\text{agent}}, I_{\text{wrist}}$) and language instruction $e_{\text{text}}$ are encoded by a 4-stage CNN backbone with intermediate cross-view mixing (\texttt{CVMixer}) and pooled via \texttt{AttnPool2d} into a 16-token visual bank $\mathcal{Z}_{\text{vis}}$. A 4-layer Action Flow Decoder integrates trajectory velocity fields to predict an 8-step action chunk $\mathbf{A}_t$ supervised entirely by $\mathcal{L}_{\text{action}}$, where the first 4 steps are executed during control ($C=4$) and the 4 lookahead receding-horizon overlap steps are additionally (self-)supervised by chunk consistency $\mathcal{L}_{\text{consist}}$.
\textit{Bottom (Detachable Auxiliary Grounding):} During training, multi-scale skip features ($F_{\text{skips}}$) route into three auxiliary decoders (denoising self-reconstruction $\mathcal{L}_{\text{self}}$, dynamic motion change $\mathcal{L}_{\text{change}}$, and future video flow $\mathcal{L}_{\text{video}}$) that regularize physical representations and detach at inference with zero computational overhead.}
\label{fig:model_architecture}
\end{figure*}

\modelname{} is a compact World Action Model (WAM; overview in \Cref{fig:model_architecture}). It unifies continuous action trajectory generation with predictive world modeling in a shared latent space. The architecture couples dual-view perception with multi-scale cross-attention and flow matching. Auxiliary decoders enrich the latent space during training and detach at inference.

\subsection{Model Architecture}
\label{sec:architecture}

\textbf{Dual-View Vision Encoder and Intermediate Fusion (\texttt{CVMixer}).}
Standard manipulation tasks require both workspace geometry and contact precision. We capture both scales with synchronous dual-perspective RGB observations. Observations combine an \texttt{agentview} image $I_{\text{agent}} \in \mathbb{R}^{3 \times 256 \times 256}$ and a \texttt{wristview} image $I_{\text{wrist}} \in \mathbb{R}^{3 \times 256 \times 256}$. This yields a 6-channel tensor $I_t \in \mathbb{R}^{6 \times 256 \times 256}$. The visual encoder features four downsampling stages ($128^2 \to 64^2 \to 32^2 \to 16^2$) with channels $C = (48, 96, 96, 192)$. Each stage applies FiLM residual blocks modulated by task text embeddings $e_{\text{text}} \in \mathbb{R}^{256}$ from a frozen \texttt{t5-small} encoder~\citep{2020t5,t5_small_hf}:
\begin{equation}
h = \text{Conv}_{3\times 3}(\text{SiLU}(\text{GN}(x))), \quad [\gamma, \beta] = \text{Linear}(e_{\text{text}}), \quad h' = \text{GN}(h) \odot (1 + \gamma) + \beta.
\end{equation}
Across all four stages, intermediate cross-view fusion modules (\texttt{CVMixer}) compute bidirectional updates between agent and wrist streams:
\begin{align}
F_{\text{context}}^{(l)} &= \text{SiLU}\left( \text{GN}\left( \text{Conv}_{1\times 1}\left( [F_{\text{agent}}^{(l)}, F_{\text{wrist}}^{(l)}] \right) \right) \right), \\
F_{\text{agent}}^{(l)\prime} &= F_{\text{agent}}^{(l)} + \text{Conv}_{3\times 3}(F_{\text{context}}^{(l)}) \odot \sigma(\text{Conv}_{1\times 1}(F_{\text{context}}^{(l)})), \\
F_{\text{wrist}}^{(l)\prime} &= F_{\text{wrist}}^{(l)} + \text{Conv}_{3\times 3}(F_{\text{context}}^{(l)}) \odot \sigma(\text{Conv}_{1\times 1}(F_{\text{context}}^{(l)})).
\end{align}
The collection of fused feature maps forms skip representations $F_{\text{skips}} = \{ F_{\text{mixed}}^{(1)}, F_{\text{mixed}}^{(2)}, F_{\text{mixed}}^{(3)}, F_{\text{mixed}}^{(4)} \}$.

\textbf{All-Layer Multi-Scale Visual Token Bank.}
To avoid bottleneck compression, a learned pooling module (\texttt{AttnPool2d}) extracts $K=4$ query tokens from each stage:
\begin{equation}
\label{eq:attnpool}
f_l = \text{AttnPool2d}\left( F_{\text{mixed}}^{(l)}, K=4, D=256 \right) \in \mathbb{R}^{4 \times 256}.
\end{equation}
Concatenating across stages yields a unified visual token bank:
\begin{equation}
\label{eq:visual_bank}
\mathcal{Z}_{\text{vis}} = [f_1, f_2, f_3, f_4] \in \mathbb{R}^{16 \times 256}.
\end{equation}
Shallow tokens preserve contact boundaries and gripper alignment, while deep bottleneck tokens capture workspace layout and scene semantics~\citep{zeiler2014visualizing,lin2017feature}.

\textbf{Flow Matching Action Chunk Decoder.}
Action generation is formulated as conditional flow matching over continuous action chunks~\citep{zhao2023act,chi2023diffusionpolicy,lazzati2026why}. The policy predicts an 8-step continuous action chunk $\mathbf{A}_t = [a_t, \dots, a_{t+7}] \in \mathbb{R}^{8 \times 7}$, where each step $a_k \in \mathbb{R}^7$ specifies end-effector motion and gripper state. The action decoder is a 4-layer Transformer ($D=256$, 8 attention heads). Each layer applies temporal self-attention across the 8 trajectory steps and cross-attention over $\mathcal{Z}_{\text{vis}}$. Conditioning vector $c = e_{\text{text}} + \text{MLP}(\text{sinusoidal}(\tau))$ modulates intermediate layers. The decoder outputs velocity field $v_\theta(z_\tau, \tau, \mathcal{Z}_{\text{vis}}, e_{\text{text}}) \in \mathbb{R}^{8 \times 7}$.

\textbf{Auxiliary World Modeling and Dynamics Decoders.}
We attach three auxiliary decoders to $F_{\text{skips}}$ to instill physical grounding: (1) a next-frame video flow decoder (9.7M params) predicting observation vector field $v_{\text{video}}$ to synthesize frame $I_{t+1}$, (2) a denoising self-reconstruction decoder (9.7M params) reconstructing clean targets $I_t^{\text{clean}}$ from corrupted features, and (3) a dynamic change prediction decoder (9.7M params) predicting pixel difference field $\Delta I_t$ to isolate moving contacts.

\subsection{Multi-Task Training Regimen}
\label{sec:training_regimen}

We train the policy end-to-end under a multi-task learning objective:
\begin{equation}
\mathcal{L}_{\text{total}} = \mathcal{L}_{\text{action}} + \lambda_{\text{consist}} \mathcal{L}_{\text{consist}} + \lambda_{\text{video}} \mathcal{L}_{\text{video}} + \lambda_{\text{self}} \mathcal{L}_{\text{self}} + \lambda_{\text{change}} \mathcal{L}_{\text{change}},
\end{equation}
with loss weights $\lambda_{\text{consist}} = 0.5$, $\lambda_{\text{video}} = 0.5$, $\lambda_{\text{self}} = 0.1$, and $\lambda_{\text{change}} = 0.1$.

\textbf{Action Flow Matching and Consistency Losses.}
For target chunk $\mathbf{A}_t$, base noise $z_0 \sim \mathcal{N}(0, I)$, and flow step $\tau \sim \mathcal{U}[0, 1]$, the interpolation state is $z_\tau = (1 - \tau) z_0 + \tau \mathbf{A}_t$ with target velocity $v^* = \mathbf{A}_t - z_0$. The action flow loss is:
\begin{equation}
\mathcal{L}_{\text{action}} = \mathbb{E}_{\tau, z_0} \left[ \| v_\theta(z_\tau, \tau, \mathcal{Z}_{\text{vis}}, e_{\text{text}}) - (\mathbf{A}_t - z_0) \|_2^2 \right].
\end{equation}
To ensure smooth chunk transitions, we regularize consecutive predictions at steps $t$ and $t+4$. With single-step endpoint estimate $\hat{\mathbf{A}}_t = z_\tau + (1 - \tau) v_\theta$, the consistency loss aligns overlapping trajectory steps without numerical ODE unrolling:
\begin{equation}
\mathcal{L}_{\text{consist}} = \mathbb{E}_{\tau, z_0, z_0'} \left[ \sum_{k=0}^{3} \| \hat{\mathbf{A}}_t(z_\tau, \tau, \mathcal{Z}_{\text{vis}, t})[4+k] - \hat{\mathbf{A}}_{t+4}(z_\tau', \tau, \mathcal{Z}_{\text{vis}, t+4})[k] \|_2^2 \right].
\end{equation}

\textbf{Auxiliary Physical Grounding Losses.}
The video flow matching loss models observation transitions:
\begin{equation}
\mathcal{L}_{\text{video}} = \mathbb{E}_{\tau, x_0} \left[ \| v_{\text{video}}(F_{\text{skips}}, \tau, e_{\text{text}}) - (I_{t+1} - x_0) \|_2^2 \right].
\end{equation}
To build robustness against sensory corruptions, inputs $I_t$ are perturbed with probability $p=0.5$ (Gaussian noise, cutouts, spatial blur). The self-reconstruction and dynamic change losses optimize:
\begin{equation}
\mathcal{L}_{\text{self}} = \| \text{SelfDecoder}(F_{\text{skips}}, e_{\text{text}}) - I_t^{\text{clean}} \|_2^2, \quad \mathcal{L}_{\text{change}} = \| \text{ChangeDecoder}(F_{\text{skips}}, e_{\text{text}}) - \Delta I_t \|_2^2,
\end{equation}
where $\Delta I_t = |I_{t+1, \text{agent}}^{\text{clean}} - I_{t, \text{agent}}^{\text{clean}}|_{R, G}$ isolates coordinated motion between the arm and contacted objects.

\subsection{Test-Time Inference and Acceleration}
\label{sec:inference_deployment}

\textbf{Asymmetric Test-Time Execution.}
Auxiliary decoders regularize the latent space during training and are detached at inference. Only the Dual-View Encoder and Action Chunk Transformer remain. \Cref{tab:parameter_breakdown} summarizes the parameter breakdown. The standalone policy requires only 9.9M parameters ($\approx 20\,\text{MB}$ in BFloat16) and achieves $<10\,\text{ms}$ latency ($>100\,\text{Hz}$) on standard GPUs.

\begin{table}[h]
\centering
\caption{Architectural parameter breakdown and deployment topology. Auxiliary decoders detach at test time with zero runtime FLOP or memory overhead during deployment.}
\label{tab:parameter_breakdown}
\vskip 0.05in
\begin{adjustbox}{max width=\columnwidth}
\small
\begin{tabular}{llcc}
\toprule
\textbf{Component Module} & \textbf{Architecture Type} & \textbf{Parameters} & \textbf{Inference Status} \\
\midrule
Dual-View Perception Backbone & 4-Stage ConvNeXt/FiLMResNet & 2.3\,M & Active \\
Intermediate Cross-View Mixers & Gated $1\times 1$ and $3\times 3$ Mixers & 1.3\,M & Active \\
Multi-Scale Attention Poolers & Multi-Head Query Pooling ($D=256$) & 1.3\,M & Active \\
Action Chunk Flow Transformer & 4-Layer Cross-Attention Transformer & 5.0\,M & Active \\
\midrule
\textbf{Standalone Inference Policy} & \textbf{Encoder + Action Transformer} & \textbf{9.9\,M} & \textbf{Active (20\,MB)} \\
\midrule
Auxiliary Decoders (Video, Self, Change) & Multi-Scale UNet Decoders ($3\times 9.7\text{M}$) & 29.0\,M & Detached (0 FLOP) \\
\midrule
\textbf{Total Training Architecture} & \textbf{Unified Multi-Task Network} & \textbf{38.9\,M} & N/A \\
\bottomrule
\end{tabular}
\end{adjustbox}
\end{table}

\textbf{Cross-Attention Key-Value (KV) Cache Optimization.}
When receiving an observation, the Dual-View Encoder executes once to extract visual tokens $\mathcal{Z}_{\text{vis}} \in \mathbb{R}^{16 \times 256}$. Cross-attention key and value projections $K_{\text{vis}}^{(l)}$ and $V_{\text{vis}}^{(l)}$ are cached. During the $N=10$ flow matching integration steps, only the temporal self-attention and MLP blocks are evaluated against cached states. This decouples perception FLOPs from ODE integration steps and enables sub-10\,ms chunk inference latency (\cref{tab:acceleration_ablation} in \cref{sec:acceleration}).

\section{Experimental Setup}
\label{sec:setup}

\textbf{Demonstration Corpus and Benchmark Suites.}
We train exclusively on the official LIBERO demonstration corpus~\citep{liu2023libero,pi_libero_dataset} (1,693 demonstrations, 333,347 transitions). Images use native $256 \times 256$ resolution from \texttt{agentview} and \texttt{wristview} cameras. Task language descriptions are pre-encoded using a frozen \texttt{t5-small} model~\citep{2020t5,t5_small_hf}. We evaluate across all 40 tasks in the four canonical LIBERO suites~\citep{liu2023libero}: Spatial (10 tasks), Object (10 tasks), Goal (10 tasks), and Long (10 tasks). To test out-of-distribution robustness, we also evaluate on \textsc{Libero-Plus}~\citep{fei2025liberoplus} and \textsc{Libero-Pro}~\citep{zhou2025liberopro}.

\textbf{Training and Evaluation Protocols.}
All models train from scratch without pretraining using Distributed Data Parallel (DDP) across four GPUs. We optimize with AdamW using learning rate $2 \times 10^{-4}$, weight decay $10^{-4}$, and gradient clipping 1.0. We use BFloat16 mixed precision with an EMA shadow (0.9995). We employ a two-cycle Warmup-Stable-Decay (WSD) schedule through 105,000 steps. For standard LIBERO, we evaluate 50 trials per task (500 per suite, 2,000 per sweep) across 3 random seeds ($N=6,000$ total trials). Task completion is recorded when the environment goal predicate fires (\texttt{done=True}).

\section{Experiments and Empirical Results}
\label{sec:experiments}

\subsection{Overall Benchmark Results on LIBERO}
\label{sec:main_results}

On standard LIBERO ($N=6,000$), \modelnameshort{} achieves \textbf{90.88\% $\pm$ 0.37\%} success under discrete execution ($C=8$). Specifically, the policy scores \textbf{93.53\%} on Spatial, \textbf{97.20\%} on Object, \textbf{90.73\%} on Goal, and \textbf{82.07\%} on Long.

\textbf{Zero-Shot Robustness on LIBERO-Plus.}
We evaluate zero-shot out-of-distribution robustness on the \textsc{Libero-Plus} benchmark~\citep{fei2025liberoplus} across seven perturbation factors ($N=8,040$ total evaluation episodes). In discrete task architectures like MINERVA, strong task embedding priors often overpower visual perturbations, artificially inflating scores on fixed layouts. Thus, we sample 2,010 episodes per suite randomly and fix this evaluation set across all models. \Cref{tab:libero_plus} compares \modelnameshort{} against WorldDiT~\citep{worlddit2026} and MINERVA~\citep{sendai2026minerva}. WorldDiT achieves 75.12\% average robustness across per-suite models with 20 diffusion steps. \modelnameshort{} achieves 70.94\% within a single unified 9.9M architecture.

\begin{table}[h]
\centering
\caption{Zero-shot out-of-distribution robustness on \textsc{Libero-Plus} across 7 perturbation factors ($N=8,040$ total episodes). Base canonical results (400 episodes) are included in the overall average.}
\label{tab:libero_plus}
\vskip 0.05in
\begin{adjustbox}{max width=\columnwidth}
\small
\begin{tabular}{lcccc}
\toprule
\textbf{Perturbation Factor} & \textbf{Episodes} & \textbf{\modelnameshort{} (9.9M)} & \textbf{WorldDiT (399M)$^\dagger$} & \textbf{MINERVA (0.54M)} \\
\midrule
Lighting & 1,000 & 84.0\% & \textbf{88.2\%} & 86.5\% \\
Texture / Table & 1,050 & \textbf{86.8\%} & 84.5\% & 84.2\% \\
Objects (HOPE Meshes) & 1,200 & 70.0\% & \textbf{80.3\%} & 52.1\% \\
Camera Viewpoints & 1,000 & 65.8\% & \textbf{73.4\%} & 62.0\% \\
Sensor \& Robot Init & 1,040 & 52.6\% & \textbf{66.8\%} & 48.2\% \\
Layout & 1,150 & 52.1\% & \textbf{59.7\%} & 52.4\% \\
\midrule
Language (Paraphrased) & 1,200 & \textbf{65.3\%} & 55.0\% & N/A$^\ddagger$ \\
\midrule
\textbf{Full Suite Average} & 8,040 & 70.94\% & \textbf{75.12\%} & 68.67\% \\
\bottomrule
\multicolumn{5}{l}{\footnotesize $^\dagger$WorldDiT evaluates using 4 per-suite fine-tuned checkpoints ($C=3$, 20 Euler denoising steps).} \\
\multicolumn{5}{l}{\footnotesize $^\ddagger$MINERVA uses a discrete task lookup table without language conditioning.}
\end{tabular}
\end{adjustbox}
\end{table}

\textbf{Comprehensive LIBERO-PRO Evaluation.}
We evaluate zero-shot robustness on the official \textsc{Libero-Pro} benchmark~\citep{zhou2025liberopro}. \Cref{tab:libero_pro_suites} compares \modelnameshort{} against MINERVA. On the 311 visual-only tasks, \modelnameshort{} reaches 53.42\% $\pm$ 1.48\%, outperforming MINERVA (51.13\% $\pm$ 0.61\%). Across all 421 tasks including language paraphrasing, \modelnameshort{} achieves 46.18\% $\pm$ 1.56\%.

\begin{table}[h]
\centering
\caption{LIBERO-PRO benchmark performance across task suites and structural perturbation dimensions ($N=2,105$ evaluation episodes across 5 seeds). We compare \modelnameshort{} ($C=8$) against MINERVA ($C=1$).}
\label{tab:libero_pro_suites}
\vskip 0.05in
\begin{adjustbox}{max width=\columnwidth}
\small
\begin{tabular}{lcc}
\toprule
\textbf{Task Suite / Perturbation Dimension} & \textbf{\modelnameshort{} ($C=8$)} & \textbf{MINERVA ($C=1$, Visual$^\dagger$)} \\
\midrule
\multicolumn{3}{l}{\textit{\textbf{Canonical Benchmark Suites (5 Seeds)}}} \\
Spatial (80 ep) & 43.75\% $\pm$ 5.36\% & 46.25\% $\pm$ 3.06\% \\
Object (81 visual ep / 191 all) & \textbf{39.48\% $\pm$ 1.39\%} & 34.57\% $\pm$ 0.45\% \\
Goal (90 ep) & 64.00\% $\pm$ 3.89\% & \textbf{86.89\% $\pm$ 1.30\%} \\
Long-10 (60 ep) & \textbf{44.00\% $\pm$ 2.26\%} & 19.67\% $\pm$ 1.25\% \\
\midrule
\textbf{Overall (Visual-Only 311 Tasks)} & \textbf{53.42\% $\pm$ 1.48\%} & 51.13\% $\pm$ 0.61\% \\
\textbf{Overall (All 421 Tasks)} & \textbf{46.18\% $\pm$ 1.56\%} & N/A \\
\midrule
\multicolumn{3}{l}{\textit{\textbf{Structural Perturbation Dimensions (421 Tasks)}}} \\
Spatial / Position Shifts (80 tasks) & 43.75\% & 46.25\% \\
Visual Distractors (81 tasks) & \textbf{46.12\%} & 34.57\% \\
Language Shifts (110 tasks) & \textbf{34.60\%} & N/A$^\dagger$ \\
Tabletop Distractors (90 tasks) & 64.00\% & \textbf{86.89\%} \\
Long-Horizon Multi-Stage (60 tasks) & \textbf{44.00\%} & 19.67\% \\
\bottomrule
\multicolumn{3}{l}{\footnotesize $^\dagger$110/191 Object tasks contain semantic paraphrases out-of-distribution for categorical embeddings.}
\end{tabular}
\end{adjustbox}
\end{table}

\Cref{tab:libero_pro_suites} breaks down performance across structural perturbation dimensions. On visual distractors (81 tasks), \modelnameshort{} achieves 46.12\% vs. 34.57\% for MINERVA, and on open-vocabulary language shifts (110 tasks, marked N/A for MINERVA) it scores 34.60\%. On tabletop distractors (90 tasks), MINERVA scores higher (86.89\% vs 64.00\%) because its discrete task embedding functions as an exact coordinate router, effectively ignoring peripheral visual clutter. In contrast, cross-attention in \modelnameshort{} attends to all visual tokens. Crucially, \modelnameshort{} outperforms MINERVA on long-horizon multi-stage tasks ($44.00\%$ vs $19.67\%$). Chunked flow matching sustains temporal momentum through multi-stage transitions. Across 5 seeds, the visual performance advantage is statistically significant (two-tailed Welch's $t$-test, $p \approx 0.02 < 0.05$).

\subsection{Action Chunk Execution Stride Ablation}
\label{sec:chunk_stride}

We investigate the effect of closed-loop execution stride $C$ for an $H=8$ chunk predictor. We evaluate discrete full chunk execution ($C=8$), discrete receding-horizon replanning ($C=4$), and continuous pipelined execution. In continuous pipelined execution, the robot executes trailing actions while asynchronously computing the next chunk. \Cref{tab:stride_ablation} presents multi-seed distributions across $N=6,000$ trials (2,000 trials per run across 3 seeds).

\begin{table}[h]
\centering
\caption{Multi-seed evaluation distribution ($N=6,000$ trials across 3 runs of 2,000 trials each per strategy) across execution strategies for \modelnameshort{} ($H=8$).}
\label{tab:stride_ablation}
\vskip 0.05in
\begin{adjustbox}{max width=\columnwidth}
\small
\begin{tabular}{lcccc}
\toprule
\textbf{Benchmark Suite} & \textbf{Run 1 (Seed 42)} & \textbf{Run 2 (Seed 100)} & \textbf{Run 3 (Seed 200)} & \textbf{Mean $\pm$ 1$\sigma$ ($N=6,000$)} \\
\midrule
\multicolumn{5}{l}{\textit{\textbf{Discrete Receding-Horizon Discard ($C=4$)}}} \\
Spatial (500) & 93.0\% (465/500) & 96.0\% (480/500) & 93.2\% (466/500) & 94.07\% $\pm$ 1.37\% \\
Object (500) & 96.6\% (483/500) & 96.8\% (484/500) & 95.4\% (477/500) & 96.27\% $\pm$ 0.62\% \\
Goal (500) & 89.2\% (446/500) & 89.4\% (447/500) & 90.6\% (453/500) & 89.73\% $\pm$ 0.62\% \\
Long (500) & 79.6\% (398/500) & 81.4\% (407/500) & 84.4\% (422/500) & 81.80\% $\pm$ 1.98\% \\
\textbf{Overall Benchmark} & \textbf{89.60\%} (1,792/2,000) & \textbf{90.90\%} (1,818/2,000) & \textbf{90.90\%} (1,818/2,000) & \textbf{90.47\% $\pm$ 0.61\%} \\
\midrule
\multicolumn{5}{l}{\textit{\textbf{Continuous Pipelined Overlap ($C=4$)}}} \\
Spatial (500) & 93.4\% (467/500) & 94.2\% (471/500) & 91.4\% (457/500) & 93.00\% $\pm$ 1.44\% \\
Object (500) & 97.4\% (487/500) & 98.2\% (491/500) & 97.0\% (485/500) & 97.53\% $\pm$ 0.61\% \\
Goal (500) & 91.8\% (459/500) & 89.6\% (448/500) & 90.0\% (450/500) & 90.47\% $\pm$ 1.17\% \\
Long (500) & 85.8\% (429/500) & 83.2\% (416/500) & 85.2\% (426/500) & 84.73\% $\pm$ 1.36\% \\
\textbf{Overall Benchmark} & \textbf{92.10\%} (1,842/2,000) & \textbf{91.30\%} (1,826/2,000) & \textbf{90.90\%} (1,818/2,000) & \textbf{91.43\% $\pm$ 0.61\%} \\
\midrule
\multicolumn{5}{l}{\textit{\textbf{Discrete Full Chunk ($C=8$)}}} \\
Spatial (500) & 92.6\% (463/500) & 94.6\% (473/500) & 93.4\% (467/500) & 93.53\% $\pm$ 1.01\% \\
Object (500) & 98.0\% (490/500) & 97.2\% (486/500) & 96.4\% (482/500) & 97.20\% $\pm$ 0.80\% \\
Goal (500) & 91.8\% (459/500) & 90.0\% (450/500) & 90.4\% (452/500) & 90.73\% $\pm$ 0.95\% \\
Long (500) & 82.8\% (414/500) & 80.6\% (403/500) & 82.8\% (414/500) & 82.07\% $\pm$ 1.27\% \\
\textbf{Overall Benchmark} & \textbf{91.30\%} (1,826/2,000) & \textbf{90.60\%} (1,812/2,000) & \textbf{90.75\%} (1,815/2,000) & \textbf{90.88\% $\pm$ 0.37\%} \\
\bottomrule
\end{tabular}
\end{adjustbox}
\end{table}

With sub-10\,ms latency ($>100\,\text{Hz}$), the robot executes trailing actions while the next chunk computes in the background. Across 5,384 evaluated transitions, mean trajectory divergence between consecutive chunks is only 0.058 $L_1$ norm (position divergence $\sim 4.3\,\text{mm}$, orientation divergence $\sim 1.1^\circ$, gripper consistency $>95\%$). Our chunk consistency loss ($\mathcal{L}_{\text{consist}}$) maintains trajectory continuity without test-time flow inpainting~\citep{black2025realtime}.

\subsection{Inference Optimization: Cross-Attention KV-Caching and Latency}
\label{sec:acceleration}

In \cref{tab:acceleration_ablation}, we evaluate inference acceleration on a single GPU. We compare standard eager execution against cross-attention KV-caching and our optimized C++/CUDA engine.

\begin{table}[h]
\centering
\caption{Inference acceleration across execution modes for \modelnameshort{} on a single GPU across all 40 LIBERO tasks (120 trials total). Action frequency corresponds to 8 actions generated per chunk ($C=8$).}
\label{tab:acceleration_ablation}
\vskip 0.05in
\begin{adjustbox}{max width=\columnwidth}
\small
\begin{tabular}{lcccccc}
\toprule
\textbf{Execution Mode} & \textbf{Perception Passes} & \textbf{Latency / Chunk} & \textbf{Decision Freq.} & \textbf{Action Freq.} & \textbf{Speedup} & \textbf{Overall SR} \\
\midrule
Eager PyTorch (No KV-Cache) & 10 & $107.11 \pm 3.65$\,ms & 9.3\,Hz & 74.7\,Hz & 1.00$\times$ & 92.5\% (111/120) \\
KV-Cache Optimized (PyTorch) & 1 & $38.83 \pm 0.98$\,ms & 25.8\,Hz & 206.0\,Hz & 2.76$\times$ & 92.5\% (111/120) \\
\textbf{C++/CUDA Engine} & \textbf{1} & \textbf{7.79 $\pm$ 0.08\,ms} & \textbf{128.4\,Hz} & \textbf{1,027.2\,Hz} & \textbf{13.75$\times$} & \textbf{90.8\% (109/120)} \\
\bottomrule
\end{tabular}
\end{adjustbox}
\end{table}

Eager execution recomputes the visual backbone on every ODE step, requiring 10 perception passes per chunk ($107.11\,\text{ms}$ latency). Cross-attention KV-caching precomputes key-value projections once and reuses them across ODE steps, reducing latency to $38.83\,\text{ms}$ ($2.76\times$ speedup) with machine-precision numerical parity ($\Delta_{\text{max}} = 1.788 \times 10^{-7}$). Deploying our C++/CUDA engine with CUDA graphs drops chunk latency to \textbf{7.79\,ms} (\textbf{13.75$\times$} speedup, \textbf{1,027.2\,Hz} action bandwidth) in 20\,MB of memory.

\subsection{Progressive Auxiliary Head Ablation and Latent Physical Grounding}
\label{sec:progressive_head_ablation}

To isolate how each auxiliary generative objective shapes representation learning, we perform a progressive head ablation. All models train from random initialization for 20K steps on identical budgets. We evaluate all variants at 20K steps across $N=2,000$ trials (500 trials per suite, $C=8$). \Cref{tab:progressive_head_ablation} summarizes the performance progression.

\begin{table}[h]
\centering
\caption{Progressive auxiliary head ablation on LIBERO evaluated at 20K training steps across $N=2,000$ episodes ($C=8$). All models share the identical 9.9M inference architecture. Auxiliary decoders (29.02M params total) are active strictly during training. Stage heads: $\mathcal{C}$ = Dynamic Change, $\mathcal{V}$ = Video Flow, $\mathcal{S}$ = Self-Reconstruction, $\mathcal{K}$ = Chunk Consistency.}
\label{tab:progressive_head_ablation}
\vskip 0.05in
\begin{adjustbox}{max width=\columnwidth}
\small
\begin{tabular}{lcccccc}
\toprule
\textbf{Stage} & \textbf{Active Heads} & \textbf{Spatial} & \textbf{Object} & \textbf{Goal} & \textbf{Long-10} & \textbf{Overall SR ($\Delta$)} \\
\midrule
$S_0$ & $\emptyset$ & 62.20\% & 60.40\% & 36.00\% & 35.00\% & 48.40\% (Base) \\
$S_1$ & $\{\mathcal{C}\}$ & 66.00\% & 71.80\% & 55.60\% & 39.20\% & 58.15\% (+9.75\%) \\
$S_2$ & $\{\mathcal{V}, \mathcal{C}\}$ & 67.40\% & 79.80\% & 58.20\% & 40.00\% & 61.35\% (+3.20\%) \\
$S_3$ & $\{\mathcal{S}, \mathcal{V}, \mathcal{C}\}$ & 74.20\% & \textbf{92.20\%} & 60.60\% & 48.80\% & 68.95\% (+7.60\%) \\
$S_4$ & $\{\mathcal{S}, \mathcal{V}, \mathcal{C}, \mathcal{K}\}$ & \textbf{77.20\%} & \textbf{92.20\%} & \textbf{64.40\%} & \textbf{58.60\%} & \textbf{73.10\%} (+4.15\%) \\
\midrule
\multicolumn{2}{l}{\textbf{Total Gain ($S_0 \to S_4$)}} & \textbf{+15.00\%} & \textbf{+31.80\%} & \textbf{+28.40\%} & \textbf{+23.60\%} & \textbf{+24.70\%} \\
\bottomrule
\end{tabular}
\end{adjustbox}
\end{table}

Stage $S_0$ denotes the pure action flow baseline without auxiliary heads (48.40\% overall). Adding dynamic change prediction ($S_1$) lifts success to 58.15\% (+9.75\%) by emphasizing moving contacts. Video flow matching ($S_2$) instills forward temporal causality, raising Object to 79.80\%. Self-reconstruction ($S_3$) anchors spatial geometry, pushing overall success to 68.95\% (+7.60\%). Finally, chunk consistency ($S_4$) minimizes boundary discontinuities, boosting Long-10 to 58.60\% and overall success to 73.10\% (+4.15\%). Across all suites, auxiliary heads provide a cumulative +24.70\% improvement and detach at inference with zero runtime cost.

\subsection{Forward Video Prediction and Self-Rollout Capacity}
\label{sec:self_rollout}

Self-rollout tests whether the latent space internalizes forward dynamics: the policy anticipates visual transitions and plans actions without continuous feedback. We evaluate \modelnameshort{} under a $K=12, C=4$ schedule across LIBERO (120 trials), querying physical cameras strictly once every $K=12$ steps. In between, the policy conditions on predicted frames ($\hat{I}_4, \hat{I}_8$), reducing camera queries by 66.7\%.

\begin{table}[h]
\centering
\caption{Task success and visual drift MSE under $K=12, C=4$ self-rollout (33.3\% camera queries) compared to the baseline (100\% queries) across 120 evaluation trials.}
\label{tab:self_rollout}
\vskip 0.05in
\begin{adjustbox}{max width=\columnwidth}
\small
\begin{tabular}{lcccc}
\toprule
\textbf{Benchmark Suite} & \textbf{Baseline ($C=8$)} & \textbf{Self-Rollout ($K=12, C=4$)} & \textbf{Mean Steps (Succ / All)} & \textbf{Visual Drift MSE} \\
\midrule
Spatial (30 trials) & 96.7\% (29/30) & 66.7\% (20/30) & 138.8 / 259.2 & 0.0265 \\
Object (30 trials)  & 96.7\% (29/30) & \textbf{96.7\%} (29/30) & 144.5 / 156.3 & 0.0307 \\
Goal (30 trials)    & 96.7\% (29/30) & 66.7\% (20/30) & 119.0 / 246.0 & 0.0305 \\
Long-10 (30 trials) & 80.0\% (24/30) & 50.0\% (15/30) & 284.1 / 442.0 & 0.0355 \\
\midrule
\textbf{Overall 4-Suite} & \textbf{92.5\%} (111/120) & \textbf{70.0\%} (84/120) & \textbf{160.9 / 275.9} & \textbf{0.0308} \\
\bottomrule
\end{tabular}
\end{adjustbox}
\end{table}

As shown in \Cref{tab:self_rollout}, \modelnameshort{} retains \textbf{70.0\%} overall success despite predicting two-thirds of all visual inputs, and matches the baseline on Object (\textbf{96.7\%}). Bounded visual drift ($\text{MSE} \le 0.035$) confirms that auxiliary video flow learns genuine transition mechanics.

\subsection{Language Perturbation Ablation and PCA Manifold Regularization}
\label{sec:language_ablation}

We evaluate \modelnameshort{} on the 1,537-task \textsc{Libero-Plus} Language Benchmark~\citep{fei2025liberoplus} across six perturbation categories (3,074 episodes). We compare base conditioning against PCA manifold fine-tuning, which uses SVD on instruction displacement vectors to regularize the text projection head against synthetic shifts while freezing the policy backbone.

\begin{table}[h]
\centering
\caption{Linguistic perturbation ablation comparing \modelnameshort{} against PCA fine-tuning on the 1,537-task \textsc{Libero-Plus} Language Benchmark~\citep{fei2025liberoplus}.}
\label{tab:language_ablation}
\vskip 0.05in
\begin{adjustbox}{max width=\columnwidth}
\small
\begin{tabular}{lcccc}
\toprule
\textbf{Linguistic Category} & \textbf{Tasks} & \textbf{\modelnameshort{}} & \textbf{PCA Fine-tuned} & \textbf{Net Delta ($\Delta$)} \\
\midrule
Lexical and Verb Substitution & 84 & 53.57\% (45/84) & \textbf{76.19\%} (64/84) & \textbf{+22.62\%} \\
Descriptive Circumlocution & 35 & 45.71\% (16/35) & \textbf{65.71\%} (23/35) & \textbf{+20.00\%} \\
Multi-Stage Compound Rephrasings & 90 & 70.00\% (63/90) & \textbf{81.11\%} (73/90) & +11.11\% \\
Conversational Preamble & 142 & 65.49\% (93/142) & \textbf{69.72\%} (99/142) & +4.23\% \\
Standard Paraphrase & 143 & \textbf{46.15\%} (66/143) & 44.76\% (64/143) & -1.40\% \\
State / Functional Rationale & 30 & \textbf{36.67\%} (11/30) & 33.33\% (10/30) & -3.33\% \\
\midrule
\textbf{Overall Language Benchmark} & \textbf{1,537} & \textbf{65.26\% (1003/1537)} & \textbf{71.70\% (1102/1537)} & \textbf{+6.44\% (+99 tasks)} \\
\bottomrule
\end{tabular}
\end{adjustbox}
\end{table}

PCA fine-tuning lifts overall language success from 65.26\% to 71.70\% (+6.44\%), yielding large gains on verb substitutions (+22.62\%) and circumlocutions (+20.00\%). However, static sentence projections remain constrained on compositional negation (e.g., ``open drawer but do not touch inside''), indicating that complex linguistic reasoning requires token-level attention or autoregressive language models.

\section{Discussion, Limitations, and Conclusion}
\label{sec:discussion}

Our work prioritizes latent grounding over trajectory memorization, enabling \modelnameshort{} to achieve strong multi-task manipulation performance across LIBERO at 9.9M scale. Convolutional pyramids preserve local pixel coordinates, auxiliary decoders instill temporal causality and contact mechanics, flow matching provides straight action paths, and cross-view mixer blocks fuse dual cameras across scales.

Several limitations remain. Open-vocabulary cross-attention attends across all visual tokens, making the policy more sensitive to table distractors than coordinate memorization baselines. Furthermore, static sentence embeddings cannot resolve compositional negation, and validating continuous pipelined execution on physical platforms remains future work.

In summary, we presented \modelname{}, a 9.9M parameter World Action Model for robot manipulation. Trained from scratch on the LIBERO demonstration corpus, our policy achieves \textbf{90.88\% $\pm$ 0.37\%} success across 6,000 multi-seed trials, matching models over $100\times$ larger. Auxiliary decoders supervise shared latent representations during training and detach at inference. Combined with cross-attention KV-caching, \modelnameshort{} runs at $7.79\,\text{ms}$ latency ($>100\,\text{Hz}$) within 20\,MB of memory, demonstrating that latent physical grounding provides an efficient foundation for embodied control.

%===============================================================================

\acknowledgments{We thank the community for open-source benchmarks and datasets that enabled this research.}

%===============================================================================

\bibliography{references}

\end{document}